# A Jointed Feature Fusion Framework for Photoacoustic Reconstruction

Hengrong Lan, Changchun Yang, and Fei Gao*, *Member, IEEE*

*Abstract*—Photoacoustic (PA) computed tomography (PACT) reconstructs the initial pressure distribution from raw PA signals. The standard reconstruction of medical image could cause the artifacts due to interferences or ill-posed setup. Recently, deep learning has been used to reconstruct the PA image with ill-posed conditions. Most works remove the artifacts from image domain, and compensate the limited-view from dataset. In this paper, we propose a jointed feature fusion framework (JEFF-Net) based on deep learning to reconstruct the PA image using limited-view data. The cross-domain features from limited-view position-wise data and the reconstructed image are fused by a backtracked supervision. Specifically, our results could generate superior performance, whose artifacts are drastically reduced in the output compared to ground-truth (full-view reconstructed result). In this paper, a quarter position-wise data (32 channels) is fed into model, which outputs another 3-quarters-view data (96 channels). Moreover, two novel losses are designed to restrain the artifacts by sufficiently manipulating superposed data. The numerical and in-vivo results have demonstrated the superior performance of our method to reconstruct the full-view image without artifacts. Finally, quantitative evaluations show that our proposed method outperformed the ground-truth in some metrics.

## I. INTRODUCTION

As a hybrid imaging modality, photoacoustic tomography (PAT) has emerged to visualize the chromophores in biological tissue by converting absorbed optical energy into acoustic energy. It has high spatial resolution at deep penetration in tissues. Many potential applications have been explored in biomedical imaging areas, such as blood oxygen saturation (sO2) quantification for cancer diagnostics. Photoacoustic computed tomography (PACT) possesses high temporal resolution by reconstructing a photoacoustic (PA) image with single-shot pulsed laser light and provides potential preclinical and clinical prospects in thyroid cancer, breast cancer diagnostics, and small animal whole-body imaging [1-3]. Standard reconstruction algorithm, e.g. delay-and-sum (DAS), is widely used to rebuild PA image with high frame rate. However, ill-posed conditions, e.g. limited-view and limited elements, could cause poor quality with blurry image and artifacts. Many studies have improved the reconstruction methods to address these issues to some extent [4, 5]. These methods improved the quality of PA image paid by increasing the computational complexity of reconstruction.

Recently, deep learning (DL) has emerged to reconstruct the PA image [6, 7]. Specially, convolutional neural networks (CNN) have been a great success in the computer vision area. DL enables PAT reconstruction in both image and signal domains. In image domain, a straightforward way of applying DL is to reduce image artifacts as a post-processing step [8-10]. For instance, Neda Davoudi et al. used a U-net for efficient recovery of image quality from sparse data [11]. Also, DL directly learns the map from PA signals to PA image, which could contain a complex physical procedure [12, 13]. Derek Allman [14] used a CNN to beamform the raw data to remove the reflection artifact. However, DL has difficulty in learning the process from signal to image for complex objects, which is susceptible to noise interference for PA signal. In signal domain, DL is used to recover the bandwidth of PA signals and improve the signal-to-noise ratio (SNR) of PA signals [15, 16]. Then, a higher quality image can be rebuilt by standard reconstructed method. Some frameworks extract different features by combined signal and image instead of single domain. In [17, 18], the authors proposed a multi-input reconstruction framework by combining signal and image inputs. Steven Guan et al. used pixel-wise delayed data as input of CNN, which includes more positional information [19]. Meanwhile, MinWoo Kim et al. converted raw data into a 3-D array, where additional positional information is considered [20]. In addition, DL inspires iterative reconstruction methods to simplify the adjustment and repeating optimization for the inverse problem by learning the regularization and some parts of the

This paragraph of the first footnote will contain the date on which you submitted your paper for review. It will also contain support information, including sponsor and financial support acknowledgment. This research was funded by Natural Science Foundation of Shanghai (18ZR1425000), and National Natural Science Foundation of China (61805139).
Hengrong Lan and Changchun Yang contributed equally to this work.

Hengrong Lan is with the Hybrid Imaging System Laboratory, Shanghai Engineering Research Center of Intelligent Vision and Imaging, School of Information Science and Technology, ShanghaiTech University, Shanghai 201210, China, with Chinese Academy of Sciences, Shanghai Institute of Microsystem and Information Technology, Shanghai 200050, China, and also with University of Chinese Academy of Sciences, Beijing 100049, China (e-mail: lanhr@shanghaitech.edu.cn).
Changchun Yang is with the Hybrid Imaging System Laboratory, Shanghai Engineering Research Center of Intelligent Vision and Imaging, School of Information Science and Technology, ShanghaiTech University, Shanghai 201210, China (e-mail: yangchch@shanghaitech.edu.cn).
Fei Gao is with the Hybrid Imaging System Laboratory, Shanghai Engineering Research Center of Intelligent Vision and Imaging, School of Information Science and Technology, ShanghaiTech University, Shanghai 201210, China (*e-mail: gaofei@shanghaitech.edu.cn).

optimization procedure [21, 22].

For most artifacts removal and limited-view compensation, the artifacts are identified by taking PA image as input, which treats an image denoising task, and the missing view is also compensated by building the relationship of input and output image. Inspired by [19, 20], we make use of the position-wise data as input data and propose a jointed feature network (JEFF-Net) to reconstruct the limited-view PA image and eliminate the artifacts of reconstruction. We define the position-wise data of delay-and-sum (DAS) as a sub-image, which can be superimposed as a PA image. Meanwhile, the superimposed image provides the shape of the object, which can be fused with output position-wise data. Therefore, the common parts of the sub-images, i.e. the object, are extracted from the fused features. In this paper, we demonstrate JEFF-Net using limited-view (a quarter view) PACT data, which are fed into the model and generates delayed data of another 3 quarters positions as Fig.1 (a) shows. Furthermore, an image feature path transforms the output of the 3 quarters' positions and obtains the full-view image in every channel. Compared with the limit-view image, the output result shows superior performance. The ground-truth comes from DAS reconstruction with full-view data, which still contains distorted ingredients in the background. By this detached data arrangement, two novel losses are designed to restrain the artifacts in superposed position-wise data. Specifically, we could surpass the ground-truth in PACT by simple operation, which indicates fewer artifacts and higher contrast results than ground-truth. To go beyond supervision, for the first time, we propose beyond supervised reconstruction network (BSR-Net), which eliminates the effect of the interference in ground-truth. While BSR-Net completes the compensation task through usual supervised learning, the image is split into a state where multiple sub-images are superimposed. Meanwhile, the common parts of the sub-images, i.e. the object, are extracted through a novel residual structure. In this paper, we demonstrate BSR-Net using limited-view (a quarter view) photoacoustic computed tomography (PACT) data. Inspired by input format of data procured by isolated probes[16, 17], a quarter position-wise delayed raw data (32-channels) is fed into model and generates delayed data of another 3 quarters positions as Fig.1 (a) shows. The ground-truth come from delay-and-sum (DAS) reconstruction with full-view data, which still contains distorted ingredients in background. Specially, by virtue of this detached data arrangement, two novel losses are designed to restrain the artifacts in superposed position-wise data. We first surpass the ground-truth in PACT by designing special supervision and loss, which is embodied in less artifacts, higher contrast than ground-truth.

The numerical experiments are demonstrated to compare the standard full-view DAS reconstruction (ground-truth) with the proposed JEFF-Net. Meanwhile, we also compare the compensated view result with other deep learning models. Moreover, we perform in-vivo imaging experiments of mice abdomen to illustrate the superiority of our method. In quantitative evaluations, the results show better performance compared with ground-truth (0.667 >0.283 SSIM value).

Our contributions can be summarized as follows:

- For the first time, we introduce a DL solution to resolve the limited-view problem in PACT by feeding a quarter position-wise data.
- We try to remove the artifacts in full-view reconstructed result, which are caused by:
  ➢ Two joint feature data are used to fuse the reconstructed results, which caused the transformation of the output, i.e. 3 quarters' data, from position-wise data to reconstruct image by the backtracked supervision.
  ➢ Two novel losses are designed to mitigate interference in PA position-wise data.
- We validate our method on both synthetic and *in-vivo* PACT dataset, and compare our method with other models. We further compare our result with ground-truth in quantitative analysis.
- Finally, we share and release the code of our model and the mice dataset, using which other researchers can reproduce and train their DL model.

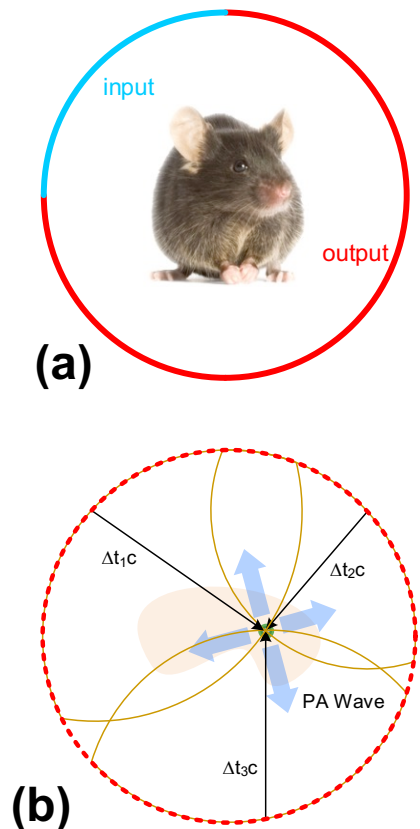

Fig. 1. (a) Illustration of the scanning setup and the input/output views of our task. (b) The propagation of PA signals and the image reconstruction principle; $\Delta t_i$ indicates the PA wave propagation time from source to detector.

## II. BACKGROUNDS

### A. Photoacoustic Computed Tomography

As a hybrid imaging modality, PAT has emerged to visualize the chromophores in biological tissue by converting absorbed optical energy into acoustic energy. It has high spatial resolution at deep penetration in tissues. Many potential applications have been explored in biomedical imaging areas, such as blood oxygen saturation (sO2) quantification for cancer diagnostics. PACT possesses high temporal resolution by reconstructing a photoacoustic (PA) image with single-shot pulsed laser light and provides potential preclinical and clinical prospects in thyroid cancer, breast cancer diagnostics, and small animal whole-body imaging [1, 2, 23, 24]. In PACT, the initial pressure is excited by the single short laser pulse, which can be expressed as [1]:

$$p_0 = \Gamma_0 \eta_{th} \mu_a F, \quad (1)$$

where $\Gamma_0$ is the Gruneisen coefficient, $\eta_{th}$ is the conversion efficiency from light to heat, $\mu_a$ is the optical absorption coefficient, and $F$ is the optical fluence. We use $x$ and $y$ to indicate the initial pressure and the received PA signals. The forward operation can be modeled as a linear operator $A$:

$$y = Ax, \quad (2)$$

which contains propagation of PA wave in the medium. The PA signals are detected by transducers as shown in Fig. 1(b). The basic idea of reconstruction is to recover x from y. For PACT, the light uniformly illuminates the whole target, which excites the PA signals simultaneously. The transducer array is used to receive the PA data at different positions. In general, the transducer with a large detection angle is desirable to receive PA signals from different directions. Several algorithms are used for PA image reconstruction, within which universal back-projection (UPB) is widely used due to less computational cost and easy implementation. In short, the basic principle of DAS can be depicted in Fig. 1(b), where PA signals are delayed to the region of interest for every channel's data based on the distance between detector and PA source.

*B. The Physical Fundamentals of Limited-view and Artifacts in PACT*

An accurate reconstruction could be maintained if the transducer fully encloses the target, and the number of elements has enough spatial density. The transducer often only accesses the PA signals from partial coverage of the tissue due to geometric restrictions. The ill-posed situation could be caused by incomplete enclosed-angle or sparse elements, which degrades image quality or lose important information. Here, we simulated different enclosed views and spatial density of transducer in Fig. 2 [25]. A full circular transducer with enough elements (e.g. 256 elements) produces a superior result as shown in Fig. 2(b). Fig. 2(c) shows that some minor artifacts will be generated if the sensors' number decreases from 256 to 128. Once the enclosed view decreases to half view, more artifacts have emerged, and the target is becoming blurry as Fig. 2(d) shows. A severe situation could happen if we further decrease the angle to a quarter-view (32 channels) as shown in Fig. 2(e). Only part of the object that is close to the sensor array can be reconstructed. Fig. 2(f) shows the result using a 128-elements linear transducer, which is polluted severely by many artifacts.

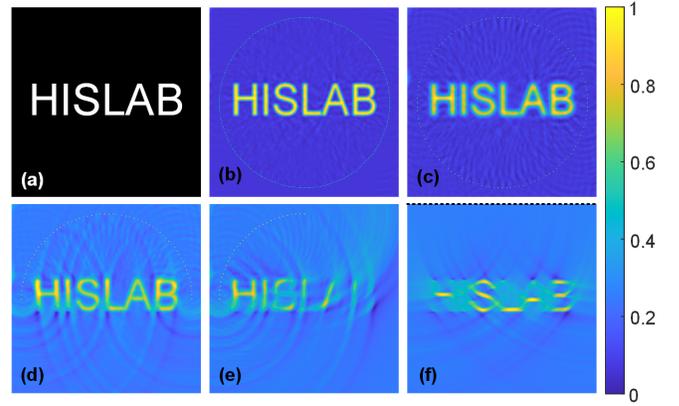

Fig. 2. Simulation results of different limited-view situations. (a) The example object. (b) reconstructed result with 256 enclosed sensors. (c) reconstructed result with 128 enclosed sensors. (d) reconstructed result with 64 half-view sensors. (e) reconstructed result with 32 quarter-view sensors. (f) reconstructed result with 128 line sensors.

Similarly, a more serious issue occurs as a result of DAS reconstruction. In standard DAS, the received PA signals are simply delayed to every pixel based on the distance from detector to pixel's position as Fig. 3(a) shows, i.e., position-wise data. Since we cannot judge the accurate orientation of PA pressure, pixels of the same radius distance are usually assigned to the same value arbitrarily. We will obtain the reconstructed PA image if we superpose all channels' delayed data as Fig. 3(b) shows. We can divide the delayed data into the data of object $d_{oj}$ and the data of artifact $d_{ar}$. Fig. 3(c) shows $d_{oj}$ of 5th-channel position-wise data, which is necessary for reconstruction. Fig. 3(d) shows $d_{ar}$ of 5th-channel position-wise data, which is the components of the artifacts, although they are small and irrelevant for all channels. $d_{ar}$ could cause severe interference when the system has a sparse number of detectors. Likewise, the DAS result fundamentally consists of $p_{oj}$ and $p_{ar}$. Generally, $d_{oj}$ and $d_{ar}$ have a similar scale of value, and $p_{oj}$ usually has a larger scale of value than $p_{ar}$. Considering the same scale of value for position-wise data, it could be more equal to both objects and artifacts if we use position-wise data as the label to supervise model, so that we can strip out an unblemished object.

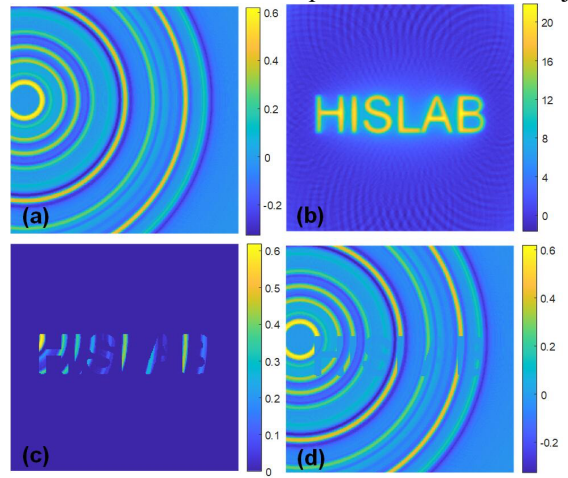

Fig. 3. Simulation results of DAS reconstruction. (a) The 5th-channel position-wise data. (b) The superposed results of 128-channel position-wise data. (c) $d_{oj}$ of 5th-channel. (d) $d_{ar}$ of 5th-channel.

## III. METHOD

To go beyond supervision, we propose a novel framework to surpass the ground-truth (reconstructed image with a full circular transducer). In this section, we use a CNN model to compensate for the limited-view of PA data and propose space-based calibration and transition module to calibrate the position-wise data. And then, to restrain the output position-wise data, we introduce an image feature path, which provides an image feature (the shape of the object) with backtracked supervision. Finally, we will propose the complete JEFF-Net with two novel losses to achieve a superior result than ground-truth.

### A. PA Position-wise Data for reconstruction

To effectively resolve the limited-view problem, the plain method expects to train an end-to-end network by feeding a limited-view result. This scheme is regarded as image repair to learn the lost information from the data with the powerful deep learning approach, which could ignore the underlying physical meaning of each channel data. Inspired by previous literature [19, 20], for the first time to our best knowledge, we introduce a limited-view compensation for position-wise data, which simply delays the raw PA signals to every pixel. In this way, the position-wise data, which contain a location relation between each channel and the respective channel, are treated as inputs of the CNN.

In our work, we demonstrate a quarter view data $x$ with 32 channels as input of our model, which can be denoted by:

$$x = [d_1(h,w), d_2(h,w), \cdots, d_{32}(h,w)]. \quad (3)$$

where $h$ and $w$ indicate height and width. The model should generate another 96 channels' data by feeding x. We express this procedure as:

$$G(x) = [d_{33}(h,w), d_{34}(h,w), \cdots, d_{128}(h,w)], \quad (4)$$

where $G(\cdot)$ denotes the DL model, and these 96 channels' data are distributed covering the remaining 270° angles. Finally, we superpose these 128 channels' data, i.e., $F(x) = \Sigma G(x) + \Sigma x$, and obtain a full-view result of DAS.

### B. Image Feature Path for Jointed Feature Fusion

In previous works, most DL-based methods used limited-view PA image to train their end-to-end model, which crudely treats the input as an incomplete image. These model could not be sensitive to small values due to differences of scale in DAS result. However, the main structure of the object can be boosted in DAS result since $p_{oj}$ is a common part for each channel. On the other hand, position-wise data distribute $p_{oj}$ and $p_{ar}$ in every channel ($p_{oj}=\Sigma d_{oj}$, $p_{ar}=\Sigma d_{ar}$), which have similar scale. We use position-wise data to equilibrate the weight between objects and artifacts of every channel.

In general, the post-processing scheme of deep learning enhances the limited-view image, which is restrained by full-view image. Practically, it is difficult to acquire the ground-truth image, so we reconstruct the full-view image with artifacts by the above operation. To improve the quality of output image, we further introduce the image feature path using the 32 channels' superimposed image as input, which guides the transformation of 96 channels' data with the backtracked supervision. We consider this branch separately, the limited-view reconstructed PA image is fed into a CNN model, obtaining the output of full-view image. This scheme is a commonly used post-processing solution to enhance the quality of PA image in ill-condition. We use the full-view DAS image as the ground-truth of this path. Moreover, we combine these two paths and add some additional losses to achieve the feature fusion (these losses will be introduced in the next section). Finally, two different features are used to reconstruct the PA image, which comes from: (1). Same weight between object and artifact of position-wise data; (2). The object with a high weight of reconstructed image.

### C. Jointed Feature Fusion Framework

As mentioned above, we integrate these structures and introduce a novel JEFF-Net to surpass the quality of ground-truth, which consists of two components as Fig. 4 shows. To fully leverage the benefit of complementary information from highly similar tasks, we proposed space-based calibration and transition module (SCTM) and two novel losses (response loss and overlay loss) to fuse the features and reconstruct the image. We design a backtracked point (BTP) in that two additional losses are used to besiege the output at the positions before and after the output, which is also called backtracked supervision. We desire that these losses can restrain $d_{oj}$ of every channel position-wise data, and obtain the artifacts and the negative object.

#### 1) Space-based calibration and transition module

The above subnetwork generates 96-channel compensation data by feeding x, and SCTM is used to replace the final layer of U-Net. We use SCTM to transfer the angle from 90° to 270°, and calibrates the relation of position-wise. SCTM has been shown in Fig. 4, which connects the encoder features to the decoder.

We design SCTM with a spatially fully-connected layer inspired by Ref. [26]. For the given feature map from the encoder, we first conduct two transformations with max pooling and average pooling. These two feature maps are further fused with grouped fully-connected layer, intended to propagate information of the corresponding position. If the input has m feature maps of size $n \times n$, the grouped fully-connected layer can decrease the number of parameters from $mn^4$ to $2mn^2$ compared to fully-connected layer. Finally, the feature map should be reshaped to $n \times n \times m$.

#### 2) JEFF-Net

We further introduce JEFF-Net to integrate the above modules. The overall architecture has been shown in Fig. 4, which can be divided into two pipelines: position-wise data

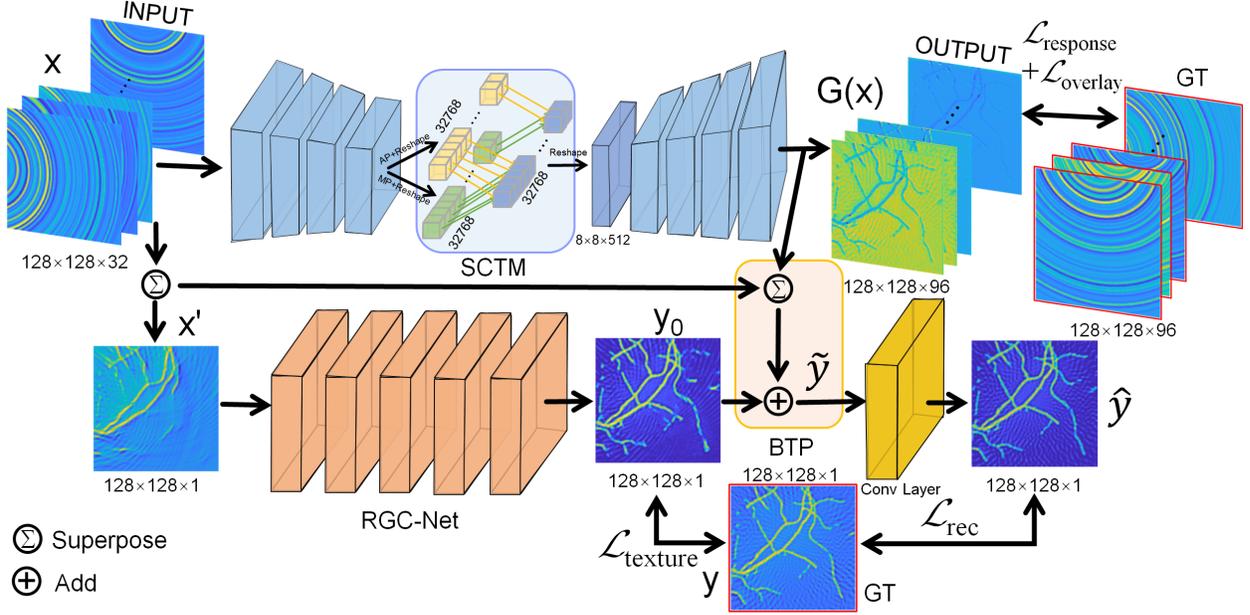

Fig. 4. The overall of proposed JEFF-Net architecture. SCTM: Space-based Calibration and Transition Module. RGC-Net: Residual Global Context subnetwork. BTP: backtracked point. GT: ground-truth. Raw data are pre-reconstructed to 32-channel position-wise data x, and superposing these 32-channel data x' as inputs of network.

compensation and limited-view image inpainting.

A U-Net [27] model is used for position-wise data compensation, which takes 32-channel position-wise data with a 90° limited view. Given input data with size 128×128×32, the first four convolutional layers and the following pooling layer (fourth layer without pooling) are used to encode the position-wise data, followed by the final layer with a convolutional layer. SCTM can connect these feature maps to decoder with 8×8×128 size. SCTM is followed by a series of five up-convolutional layers, generating 96-channel position-wise data with other 270° views. To leverage the image features from limited-view, we use a small network to extract the feature of PA image, named Residual Global Context subnetwork (RGC-Net), which consists of five residual global context layers [28]. Furthermore, we introduce a backtracked supervision before and after BTP, two losses restrain the 96 channels' data output indirectly.

To fully leverage the benefit of complementary information from highly correlated data, we bridge these two results using residual separation. By combining response loss and overlay loss for G(*x*), the artifacts in the reconstructed result can be learned, which will be described in detail below.

*3) Novel loss for position-wise data path*

We train our BSR-Net by regressing to the ground-truth content of full-view PA image. However, F(*x*) has multiple equally plausible ways to satisfy the residual relation. We propose response loss and overlay loss to handle both the artifacts and the opposite object in the output.

The compensated delay data we expect to obtain is G(*x*) in Eq. (4), which consists of additional $N_l$ (96 in our paper) channels' PA signals. That means this layer $l$ has $N_l$ feature maps, each is with size $M_l$, where $M_l$ (128×128 in our paper) is the height times the width of the feature map. Although all channels' ultrasonic sensors are spatially independently placed around the imaging target, their delayed data should have a dependent response relationship at the same position. Hence, we built a response representation that computes the correlations between the different channel's responses, where the expectation is taken over the spatial extension of the delay data. The responses between any two channels are given by the Gram matrix [29] $G^l \in \mathcal{R}^{N_l \times N_l}$, where $G^l_{ij}$ is the inner product between the vectorized delay data $i$ and $j$ in layer $l$:

$$G^l_{ij} = \sum_k F^l_{ik} F^l_{jk}. \quad (5)$$

The response loss is by minimizing the mean-squared distance between the entries of the Gram matrix from the original delay data and generated delay data ($A^l$ and $G^l$ denote their respective response representations):

$$\mathcal{L}_{response} = \frac{1}{4N_l^2 M_l^2} \sum_{i,j} \left( G^l_{ij} - A^l_{ij} \right)^2. \quad (6)$$

Although when acquiring data, a single channel records the signal sequence from its own view, the resultant image is superimposed by the delayed data of all channels. The superposition of arbitrary channels still has a certain dependence. Considering the plainest case, the contribution of the superposition of the arbitrary two views to the full views is measured here. Hence, we built an overlay representation that computes the correlations between the different channels' overlays. We propose an Overlay matrix $O^l \in \mathcal{R}^{N_l \times N_l \times M_l}$ to describe the overlays between vectorized delay data of the two probes $n$ and $n'$:

$$O^l_{nn'm} = \sum_{n,n'} F^l_{nm} + F^l_{n'm}. \qquad (7)$$

The overlay loss is by minimizing the mean-squared distance between the entries of the Overlay matrix from the original delay data and generated delay data ($P^l$ and $O^l$ denote their respective overlay representations):

$$\mathcal{L}_{overlay} = \frac{1}{4N_l^2 N_l^2 M_l^2} \sum_{n,n',m} \left(O^l_{nn'm} - P^l_{nn'm}\right)^2. \qquad (8)$$

And then, we use a texture loss to supervise the limited-view image inpainting. We apply commonly used mean square error (MSE) loss as our texture loss:

$$\mathcal{L}_{texture}(y_0) = \|y - y_0\|_F^2, \qquad (9)$$

where $F$ denotes the Frobenius norm. Furthermore, a reconstruction loss is used to optimize the residual result by minimizing the mean pixel-wise error:

$$\mathcal{L}_{rec}(\hat{y}) = \|y - \hat{y}\|_F^2. \qquad (10)$$

Finally, we define the overall loss function as follow:

$$\mathcal{L}_{overall} = \lambda_{re}\mathcal{L}_{response} + \lambda_{ov}\mathcal{L}_{overlay} + \lambda_{tex}\mathcal{L}_{texture} + \lambda_{rec}\mathcal{L}_{rec}, \qquad (11)$$

where $\lambda_{re}, \lambda_{ov}, \lambda_{tex}, \lambda_{rec}$ are hyper-parameters that decide the proportion of every loss, which have different values in different experiments.

## IV. EXPERIMENTS

In this section, we validate our method using both simulation and experimental data. Furthermore, some ablation studies are demonstrated. All deep learning methods are implemented on Pytorch [30], which is an open-source framework. The high-speed graphics computing workstation is used to train our model, which consists of four NVIDIA RTX Titan graphics cards. All experiments are described in detail below. Furthermore, the source code is available at https://github.com/chenyilan/BSR-Net.

### A. Training on Synthetic Vessels Data

We use the MATLAB toolbox k-Wave [31] to generate the synthetic dataset. The detectors surround the object evenly with 18 mm radius. The center frequency of all sensors is set as 2.5 MHz with 110% fractional bandwidth. The sound speed is 1480 m/s, and the reconstructed region is set as 26 mm × 26 mm.

We use the public fundus oculi vessel DRIVE [32] as initial pressure distribution, which should be expanded by segmentation and combination to increase the data size. Finally, we have 2800 training sets and 200 test sets. In this experiment, we use 130, 0.02, 42, 60 for $\lambda_{re}, \lambda_{ov}, \lambda_{tex}, \lambda_{rec}$, respectively.

### B. Ablation study for different sub-network comparison

In this section, we train an individual G(x) without RGC-Net. Namely, we compare different channel's position-wise data with JEFF-Net. RGC-Net indicates the conventional DL scheme to resolve the limited-view problem in image domain. Therefore, we compare three different frameworks with our JEFF-Net: 1. G(x) (above sub-network); 2. RGC-Net (nether sub-network).

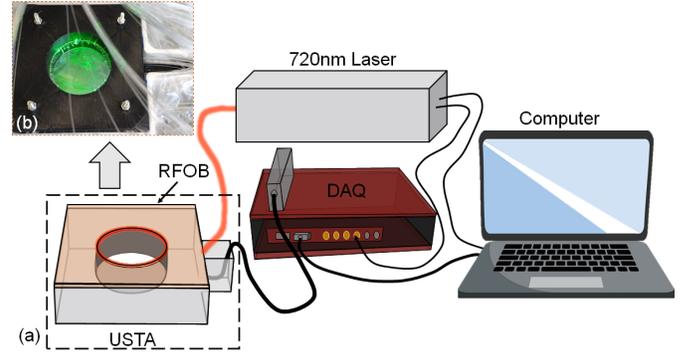

Fig. 5. (a) Schematics of the PACT system. (b) The detailed photograph of the black box region in (a). RFOB: ring-shaped fiber optics bundle; USTA: ultrasonic transducer array; DAQ: data acquisition system.

### C. Ablation study for response and overlay loss

One of the contributions in this work is to propose two novel losses. To verify them, we compare the superposed G(x) in four different cases as ablation study: remove these two losses, only use overlay loss, only use response loss, and use these two losses. Therefore, the effect of these two losses could be validated in this experiment.

### D. Training on Mice Data

Last but not least, we also verify the performance of our method on the *in vivo* data of mice abdomen. A customized PAT system (HISRing, HISLAB Inc., China) is employed to record PA signals as Fig. 5 shows, which is equipped with a 128-elements full-view ring-shaped transducer (2.5 MHz, Doppler Inc.). A pulsed laser (720 nm wavelength, 10 Hz repetition rate) is used to illuminate the object by a fiber optic bundle, which is evenly separated as a circle over the transducer as Fig. 5(b) shows, and the data sampling rate of our system is 40 MSa/s. The region of image reconstruction is 20mm×20mm.

We vertically scanned the mice using our system and obtained 1100 training sets and 116 test sets, which are available at https://ieee-dataport.org/documents/his-ring-abdomen. In this experiment, we use 250, 0.6, 30, 40 for $\lambda_{re}, \lambda_{ov}, \lambda_{tex}, \lambda_{rec}$ in Eq. (11), respectively.

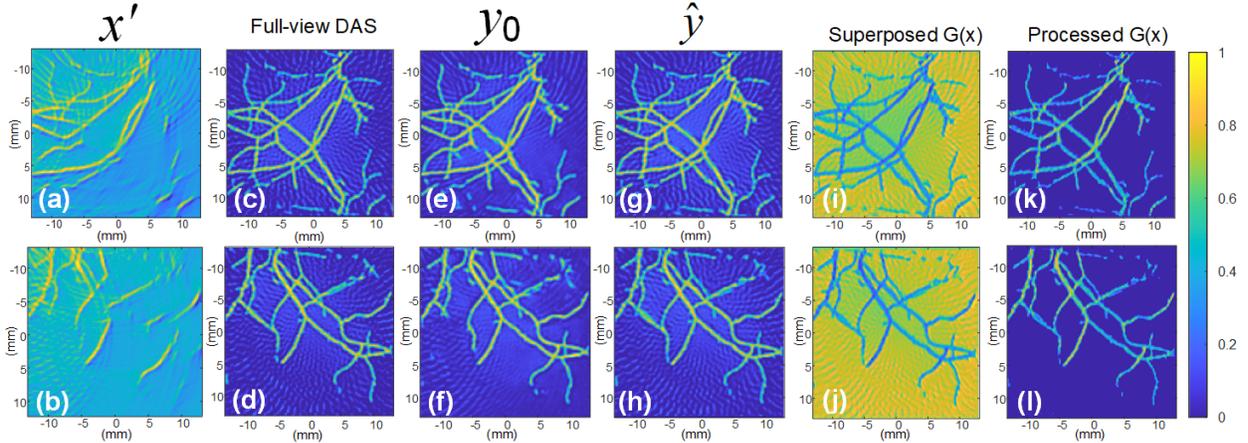

Fig. 6. Two examples of performance comparison of different results. (a,b) Limited-view DAS results. (c,d) Full-view DAS results. (e,f) $y_0$ results in JEEF-Net. (g,h) $\hat{y}$ results in JEEF-Net. (i,j) $G(x)$ superposed along channel dimension in JEEF-Net. (k,l) $G(x)$ superposed along channel dimension with the thresholding operation.

## V. RESULTS

### A. Simulation results

We show two examples of imaging results from the test set in Fig. 6, which compares the limited-view results, full-view results, and three results in the procedure of JEFF-Net. In addition, we simply process $G(x)$ and obtain the final result, which is called processed $G(x)$.

The obvious artifacts can be seen in full-view DAS from Fig. 6(c)-(d), which is the ground-truth of JEFF-Net. The whole objects are recovered from the limited-view input comparing Fig. 6(e)-(f) with Fig. 6(a)-(b). Some of the details are distorted since RGC-Net is not deep enough. The vascular structure becomes more complete after the addition of the two paths as Fig. 6(g)-(h) shows. In Fig. 6(i)-(j), the superposed 96-channel data have transformed to the artifacts and the negative object, which should be the position-wise data. $G(x)$ should be further processed to separate objects and artifacts, and here we do simple threshold processing as Fig. 6(k)-(l) shows. In fact, the target could be separated better if we explore a more advanced processing method, which will be developed in future work. For synthetic data, we have initial pressure distribution even though we do not use it in this task. Namely, we could calculate structural similarity index (SSIM) and peak signal-to-noise ratio (PSNR) to quantitatively compare these results. We use two boxplots to compare them in Fig. 7, and we only calculate processed $G(x)$ since superposed $G(x)$ is an opposite image. For average of SSIM, the processed $G(x)$ has 0.667 and the full-view DAS (the ground-truth in our work) only has 0.283. Similarly, for the average of PSNR, the processed $G(x)$ has 15.469 dB and the full-view DAS only has 14.284dB. It shows a significant superiority of our method from Fig. 7, which outperforms ground-truth (full-view DAS) in this work.

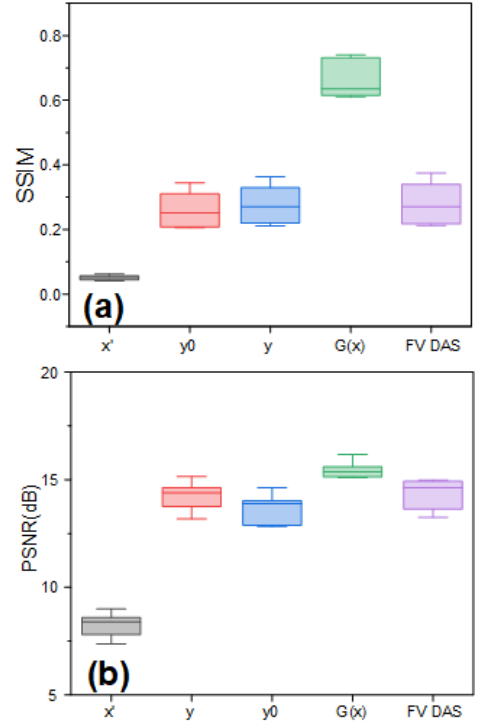

Fig. 7. The quantitative evaluations of simulated test data. (a) The boxplot of SSIM. (b) The boxplot of PSNR. $G(x)$ indicates processed $G(x)$; FV DAS: full-view DAS.

### B. Evaluation of Sub-networks

Firstly, we validate the original idea of compensating limited-view position-wise data. We only use a $G(\cdot)$ in Fig. 4 to compensate the view of PA data and plot different channel's data. In Fig. 8, we show different channel of output data and the superposed final image. Every channel of the outputs is the sensor's position-wise data at different position as shown in Fig. 8 (a)-(c), and the final image (Fig. 8 (d)) can be reconstructed by summing input and output 128 channels' data.

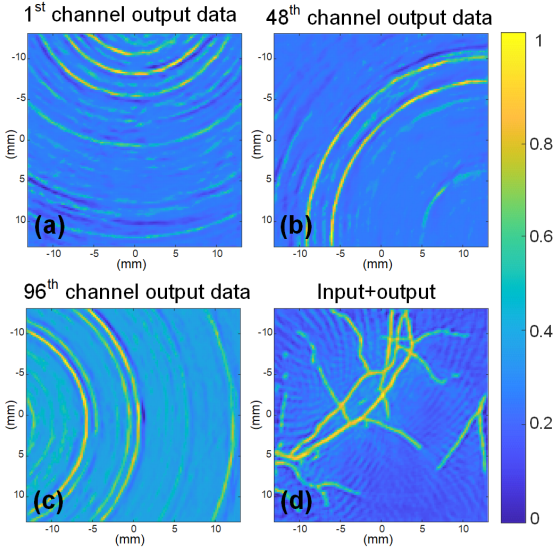

Fig. 8. The different channel of output data and the superposed data of G(·) without residual structure. (a) The 1st channel output data. (b) The 48th channel output data. (c) The 96th channel output data. (d) The sum of superposed input data and superposed output data.

Moreover, one key component of our proposed method is the different feature fusion. Fig. 9 shows the output of RGC-Net, which indicates the image post-processing scheme. Noting that some prevailing end-to-end deep learning solutions for reconstruction are often implemented by arbitrarily changing this backbone [9, 11, 33-34], which is also a comparative experiment. Fig. 9 (a) shows an obvious texture of the object, which decreases the weight of the object in the output position-wise data.

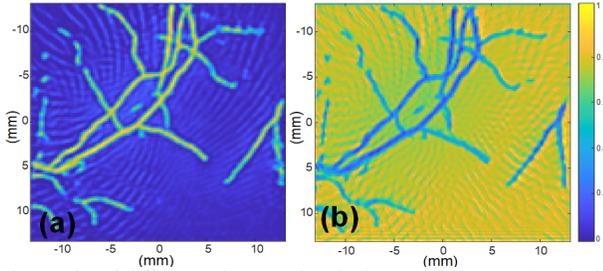

Fig. 9. Results of different sub-network; (a) The result of RGC-Net, the input is a limited-view PA image; (b). The result of superposed $G(x)$.

## C. Ablation study results

The superposed $G(x)$ of four ablation studies have been shown in Fig. 10. $G(x)$ contains both objects and artifacts with small overall value if we do not supervise $G(x)$ as Fig. 10(a) shows, which can be regarded as a supplement to $y_0$. $G(x)$ will be closer to $y$ with only one loss shown in Fig. 10(b) and (c). These two losses are designed to focus on different characteristics of position-wise data (overlay loss focuses on the relation of each channel data; response loss focuses on the common area $d_{oj}$ of each channel data). Therefore, the result of only using response loss has a higher contrast compared with that only using overlay loss. Obviously, the object and artifact can be separated into different ranges only when we use two losses simultaneously (negative objects and positive artifacts).

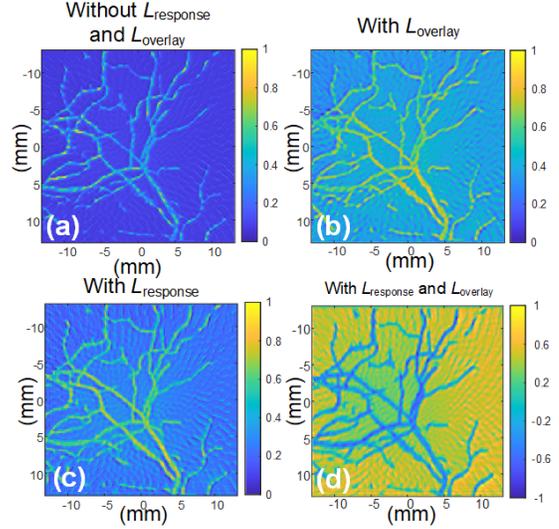

Fig. 10. The $G(x)$ results of ablation study for two novel losses. (a) The result without response loss and overlay loss. (b) The result with overlay loss. (c) The result with response loss. (d) The result with both response loss and overlay loss.

## D. In-vivo results

Lastly, we demonstrated an *in-vivo* imaging result in Fig. 11, which is the abdomen of a mouse. We also show the limited-view result with 32 detectors, the full-view result with 128 detectors, three results of JEFF-Net and the processed $G(x)$.

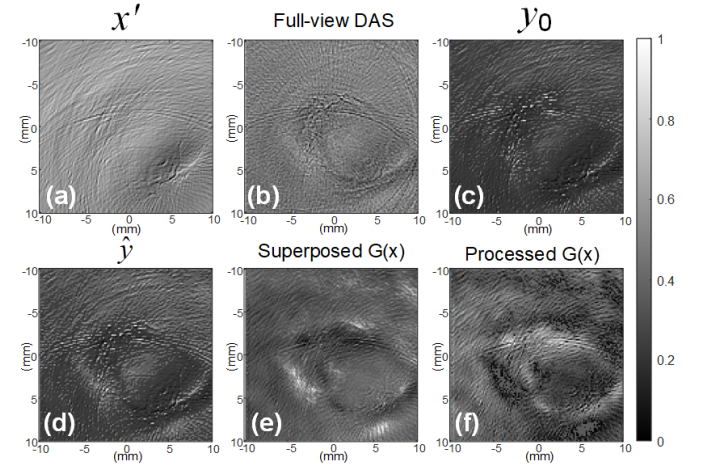

Fig. 11. The *in-vivo* results, (a) Limited-view DAS result. (b) Full-view DAS result. (c) $y_0$ result in JEFF-Net. (d) $\hat{y}$ result in JEFF-Net. (e) $G(x)$ superposed along channel dimension in JEFF-Net. (f) $G(x)$ superposed along channel dimension with thresholding operation.

Only a minor structure can be found in the limited-view result, and we even cannot recognize the outline of the object from Fig. 11(a). However, an explicit outline can be recognized from the full-view DAS result as Fig. 11(b) shows. All results from JEFF-Net show higher contrast in Fig. 11(c)-(f). The complexity of the *in-vivo* experimental condition leads to the interference of imaging objects and artifacts compared with synthetic data before, which could cause a little difference between $y_0$ and $\hat{y}$ as shown in Fig. 11(c) and (d). It is interesting that the artifacts of G(*x*) are relieved due to the residual structure in Fig. 11(e). Similarly, we further enhance Fig. 11(e) by threshold processing shown in Fig. 11(f), which clearly shows the profile of the mouse abdomen with a high contrast. Finally, we can compare these results with contrast-to-noise ratio (CNR) in Fig. 12, and it shows the best performance for our method.

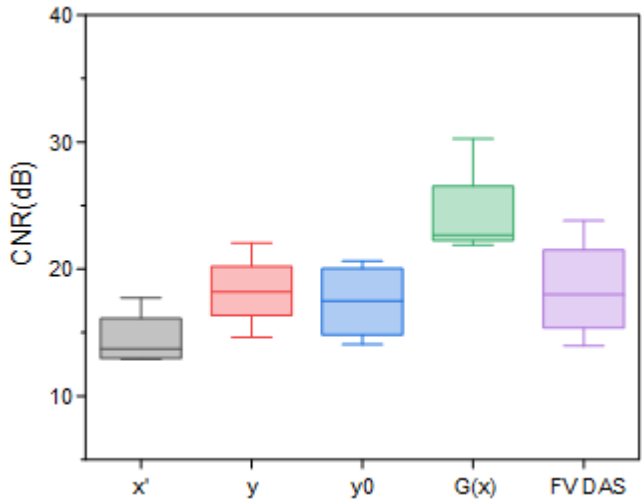

Fig. 14. The CNR of Fig. 11 *in-vivo* data. G(x) indicates processed G(x); FV DAS: full-view DAS.

## VI. Conclusion

The limited-view issue is inevitable in PACT system, which usually causes artifacts in the PA image reconstruction. In this paper, we introduce a novel framework to reconstruct the limited-view PA image: limited position-wise data are used as input of deep learning model, and generate the position-wise data of lost view. We use two different data to fuse the jointed feature for object and artifacts. To further fuse these features, a backtracked supervision is proposed, which adds redundant supervisions before and after G(*x*). This method may inspire more research fields such as image de-noising, and foreground separation. Furthermore, we proposed two novel losses to constrain the position-wise output. Therefore, we can remove the artifacts by a simple threshold processing. In our work, we propose JEFF-Net implement the proposed framework. A quarter view data is fed into the model, which outputs a group of full-view data. The numerical and *in-vivo* imaging results show that our methods have good performance compared with other models, even to ground-truth. Finally, we have also published our data and codes to facilitate other researchers for further research. It is worth noting that G(*x*) can be further used to extract more information, although we only use a threshold processing in this paper, which will be explored in future work.